\documentclass[letterpaper, 10 pt, conference]{ieeeconf}  

\IEEEoverridecommandlockouts                              

\usepackage{amssymb}
\usepackage{amsmath,amsfonts}
\usepackage{hyperref}[colorlinks,linkcolor=blue]
\usepackage{cleveref}
\usepackage{multirow}
\usepackage{makecell}
\usepackage{subcaption}
\usepackage{tcolorbox}
\usepackage[dvipsnames]{xcolor}
\definecolor{empty}{RGB}{0, 0, 0}
\definecolor{buildings}{RGB}{70, 70, 70}
\definecolor{fences}{RGB}{190, 153, 153}
\definecolor{other}{RGB}{55, 90, 80}
\definecolor{pedestrians}{RGB}{220, 20, 60}
\definecolor{poles}{RGB}{153, 153, 153}
\definecolor{roadlines}{RGB}{157, 234, 50}
\definecolor{roads}{RGB}{128, 64, 128}
\definecolor{sidewalks}{RGB}{244, 35, 232}
\definecolor{vegetation}{RGB}{107, 142, 35}
\definecolor{vehicles}{RGB}{0, 0, 142}
\definecolor{walls}{RGB}{102, 102, 156}
\definecolor{trafficsigns}{RGB}{220, 220, 0}
\definecolor{sky}{RGB}{70, 130, 180}
\definecolor{ground}{RGB}{81, 0, 81}
\definecolor{bridge}{RGB}{150, 100, 100}
\definecolor{railtrack}{RGB}{230,150,140}
\definecolor{guardrail}{RGB}{180, 165, 180}
\definecolor{trafficlight}{RGB}{250, 170, 30}
\definecolor{static}{RGB}{110, 190, 160}
\definecolor{dynamic}{RGB}{170, 120, 50}
\definecolor{water}{RGB}{45, 60, 150}
\definecolor{terrain}{RGB}{145, 170, 100}
\definecolor{unlabeled}{RGB}{160, 60, 60}

\title{\LARGE \bf
A Synthetic Benchmark for Collaborative 3D Semantic Occupancy Prediction in V2X-Enabled Autonomous Driving
}

\author{Hanlin Wu$^{1}$, Pengfei Lin$^{1}$, Ehsan Javanmardi$^{1}$, Naren Bao$^{1}$, Bo Qian$^{2}$, Hao Si$^{1}$, Manabu Tsukada$^{1}$
\thanks{$^{1}$ The School of Information Science and Technology, The University of Tokyo. 
        {\tt\small $\{$hanlinwu, linpengfei0609, ejavanmardi, naren, si-hao, mtsukada$\}$@g.ecc.u-tokyo.ac.jp}}%
\thanks{$^{2}$ National Institute of Informatics.
        {\tt\small boqian@ieee.org}}%
}

\begin{document}

\maketitle
\thispagestyle{empty}
\pagestyle{empty}

\begin{abstract}
3D semantic occupancy prediction is an emerging perception paradigm in autonomous driving, providing a voxel-level representation of both geometric details and semantic categories. However, its effectiveness is inherently constrained in single-vehicle setups by occlusions, restricted sensor range, and narrow viewpoints. To address these limitations, collaborative perception enables the exchange of complementary information, thereby enhancing the completeness and accuracy of predictions. Despite its potential, research on collaborative 3D semantic occupancy prediction is hindered by the lack of dedicated datasets. To bridge this gap, we design a high-resolution semantic voxel sensor in CARLA to produce dense and comprehensive annotations. We further develop a baseline model that performs inter-agent feature fusion via spatial alignment and attention aggregation. In addition, we establish benchmarks with varying prediction ranges designed to systematically assess the impact of spatial extent on collaborative prediction.  Experimental results demonstrate the superior performance of our baseline, with increasing gains observed as range expands. Our code is available at \hyperlink{https://github.com/tlab-wide/Co3SOP}{https://github.com/tlab-wide/Co3SOP}.
\end{abstract}

\section{Introduction}
\label{sec:intro}

Collaborative perception has emerged as a powerful strategy to enhance scene understanding in autonomous driving \cite{liu2023towards, huang2023v2x}. By sharing sensory information across multiple agents, it enables vehicles to perceive a broader environment beyond their individual field of view. This is particularly crucial in urban driving scenarios where occlusions and limited sensor range frequently hinder the perception performance of a single vehicle. Prior research \cite{xu2022v2x, hu2022where2comm, yang2024how2comm} has shown that inter-agent communication and feature fusion can significantly improve the accuracy of 3D detection, segmentation, and tracking tasks.

However, most existing collaborative perception methods employ coarse environment representations, such as 3D bounding boxes or bird's-eye-view (BEV) maps. While effective for certain tasks, these representations are insufficient for capturing the fine-grained geometry and semantics necessary for downstream reasoning and planning. Recently, 3D semantic occupancy prediction, also known as Semantic Scene Completion (SSC), has gained attention for its ability to provide detailed 3D semantic and geometric information by utilizing a fine-grain voxel-based representation \cite{xu2024survey, shi2023grid, zhang2024vision}, offering a richer and more detailed understanding of the environment.

Despite its potential, 3D semantic occupancy prediction remains largely unexplored in collaborative settings. This is due in part to the absence of dedicated datasets and benchmarks that provide voxel-level semantic ground-truth annotations under collaborative settings \cite{teufel2024collective, yazgan2024collaborative}. Existing datasets, such as SemanticKITTI \cite{behley2019iccv} and Occ3D \cite{tian2024occ3d}, lack support for multi-agent configurations and primarily rely on LiDAR point cloud of single vehicle to generate annotations. However, the inherent sparsity of LiDAR point clouds, particularly at greater distances, coupled with occlusions, sensor noise, and non-uniform point distributions, poses significant challenges to generating accurate and reliable annotations. Moreover, collaborative perception imposes higher demands on annotations, requiring temporally and spatially aligned multi-agent labels to ensure consistent supervision and effective fusion.

To address this gap, we augment an existing multi-agent dataset \cite{9812038} by replaying it in CARLA with a high-resolution semantic voxel sensor, yielding dense voxel-level annotations. The resulting dataset, which we term Co3SOP, supports the training and evaluation of collaborative 3D semantic occupancy prediction models under simulation scenarios. Additionally, we introduce a baseline model that incorporates inter-agent feature fusion with spatial alignment and sparse attention. Specifically, we apply warping-based spatial alignment to transform neighboring agents' features into the ego frame, followed by a visibility-guided sparse attention mechanism modulated by a learned confidence mask to adaptively weight contributions from each agent. Alongside the dataset and baseline, we establish benchmarks with different prediction ranges, to systematically assess how spatial distance influences the effectiveness of collaboration. 

\begin{table*}[htbp]
\renewcommand\arraystretch{1.23}
\centering
\caption{\textbf{Comparison of the existing 3D Occupancy Prediction Dataset for Autonomous Driving and our proposed Co3SOP}. Real and Sim represent the raw data is collected from real world or simulation platform. Gen means the 3D occupancy annotations are generated from raw data. C, L denote camera and LiDAR.}
\begin{center}
\scalebox{1.16}{
\begin{tabular}{c c c c c c c}
\hline
\textbf{Dataset} &\textbf{Source} &\textbf{Meta Dataset} &\textbf{Modality} &\textbf{Voxels Size} &\textbf{Resolution} &\textbf{V2X Support}\\
\hline
SemanticKITTI &  Real+Gen & KITTI & C\&L & [256, 256, 32] & [0.2, 0.2, 0.2] & - \\
KITTI-360  & Real+Gen & KITTI & C\&L & [256, 256, 32] & [0.2, 0.2, 0.2] & - \\
Occ3D-nuScenes & Real+Gen & nuScenes & C\&L& [200, 200, 16] & [0.4, 0.4, 0.4] & - \\
Occ3D-Waymo  & Real+Gen & Waymo & C\&L & [3200, 3200, 128] & [0.05, 0.05, 0.05] & - \\
SSCBench  & Real+Gen & \makecell*[c]{nuScenes\&Waymo\\\&KITTI-360} & C\&L & [256, 256, 32] & [0.2, 0.2, 0.2] & - \\
V2VSSC  & Sim+Gen & OPV2V & C\&L & [128, 128, 20] & [0.78, 0.78, 0.4] & V2V \\
Co3SOP (Ours) & Sim & OPV2V & C\&L & [1000, 1000, 70] & [0.1, 0.1, 0.1] & V2V \\
\hline
\end{tabular}
}
\end{center}
\label{tab:1}
\end{table*}

To benchmark the effectiveness of our dataset and proposed baseline model, we conduct extensive experiments on our proposed dataset, evaluating a range of state-of-the-art single-agent models as well as our collaborative baseline across all benchmark splits. The results demonstrate that collaborative perception significantly boosts semantic occupancy prediction performance, with larger collaboration ranges yielding greater improvements. In addition, we further incorporate pose noise simulation to approximate real-world uncertainty in inter-agent communication and sensing. These findings reinforce the potential of voxel-based representations in collaborative 3D scene understanding and establish Co3SOP as a foundation for future research in this direction.

\section{Related Work}

\label{sec:relatedwork}
\subsection{3D Semantic Occupancy Prediction}
\textbf{Methodologies:} 3D semantic occupancy prediction aims to provide voxel-level understanding of both geometry and semantics, and has seen rapid development across different sensing modalities. Early approaches such as SSCNet \cite{song2017semantic} adopt volumetric CNNs on RGB-D inputs. LiDAR-based methods like LMSCNet \cite{9320442}, S3CNet \cite{pmlr-v155-cheng21a}, and JS3CNet \cite{yan2021sparse} exploit the structural sparsity and accuracy of 3D point clouds to produce high-quality voxel predictions. Meanwhile, vision-based methods, including MonoScene \cite{Cao_2022_CVPR}, infer 3D semantics by lifting monocular depth and segmentation into voxel space. To overcome the limitations of single-view inputs, recent works such as VoxFormer \cite{10203337} and OccFormer \cite{Zhang_2023_ICCV} leverage multi-camera fusion and transformer-based 2D-to-3D attention mechanisms. Despite steady progress, these methods are restricted to single-agent perception, and thus remain vulnerable to occlusion, field-of-view limitations, and sensor sparsity.

\noindent\textbf{Datasets:} To facilitate the development of 3D semantic occupancy prediction, several datasets have been proposed with voxel-level annotations derived primarily from LiDAR scans. SemanticKITTI \cite{behley2019iccv, behley2021ijrr, geiger2012cvpr} is a widely adopted benchmark, providing dense semantic labels on voxelized point clouds. KITTI-360 \cite{Liao2022PAMI} extends this setup with panoramic imagery and 360° LiDAR coverage. Occ3D \cite{tian2024occ3d} introduces a voxel annotation pipeline based on Waymo and nuScenes data, offering large-scale semantic occupancy labels. SSCBench \cite{li2023sscbench} further consolidates multiple sources to unify evaluation across diverse urban scenes. While these datasets advance single-agent occupancy prediction, they do not support multi-agent collaboration. V2VSSC \cite{zhang2024v2vssc} is the first to target vehicle-to-vehicle (V2V) collaborative occupancy tasks, yet its annotations remain limited by the sparsity and occlusion inherent in LiDAR data, compromising label density and accuracy. In contrast, our proposed Co3SOP dataset offers simulation-based dense annotations specifically tailored for collaborative settings, as summarized in \cref{tab:1}.

\begin{figure*}
    \centering
    \includegraphics[width=0.98\linewidth]{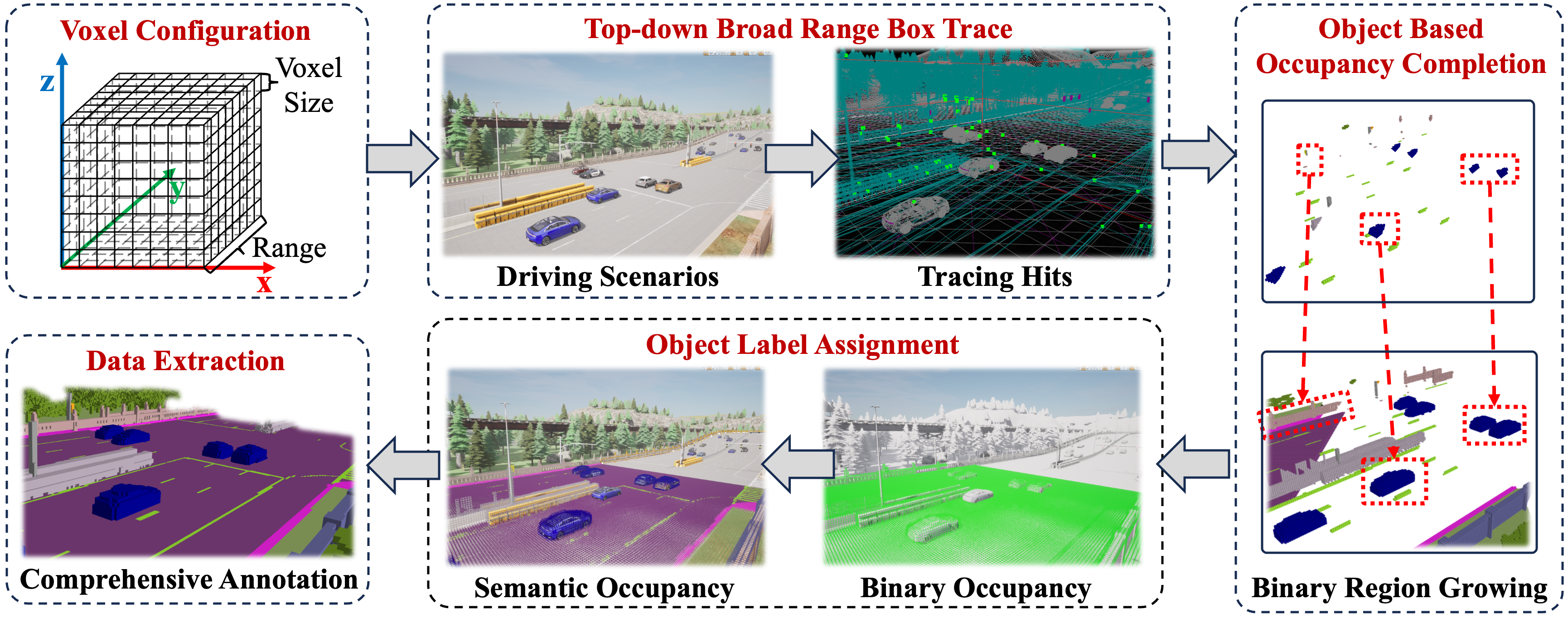}
    \captionof{figure}{Illustration of the annotation pipeline for 3D semantic voxel using a custom sensor in Carla, including voxel configuration, top-down broad range box trace, object based occupancy completion and label assignment.}
    \label{fig:voxel_sensor}
\end{figure*}

\subsection{Collaborative Perception}
\textbf{Methodologies:} Collaborative perception enhances environmental understanding by enabling vehicles to exchange complementary information. A variety of methods have been proposed to realize effective multi-agent fusion. V2VNet \cite{wang2020v2vnet} adopts graph neural networks to aggregate features from spatially distributed agents. V2X-ViT \cite{xu2022v2x} leverages vision transformers to model inter-agent interactions through attention mechanisms. To improve communication efficiency, Where2comm \cite{hu2022where2comm} identifies task-relevant regions for selective transmission, while How2comm \cite{yang2024how2comm} explores policies for adaptive communication scheduling. While effective for tasks such as object detection and segmentation, these approaches are limited to object-centric outputs and fall short in supporting fine-grain voxel-level semantic understanding.




\noindent\textbf{Datasets:} Ego-vehicle datasets such as KITTI \cite{mei2022waymo}, nuScenes \cite{caesar2020nuscenes}, and Waymo \cite{mei2022waymo} fall short in supporting collaborative perception, as they lack multi-agent scenarios and aligned multi-view annotations. To address these limitations, a range of specialized datasets have emerged \cite{teufel2024collective, yazgan2024collaborative, zimmer2024tumtraf, hao2024rcooper}. Among simulated datasets, V2X-Sim \cite{li2022v2x} models V2X interactions in CARLA and supports tasks such as detection, tracking, and BEV segmentation. OPV2V \cite{9812038}, also built in CARLA, emphasizes real-time V2V communication and provides configurable multi-agent scenarios. In real-world settings, V2V4Real \cite{xu2023v2v4real} focuses on vehicle-to-vehicle collaboration, while DAIR-V2X \cite{yu2022dair} targets vehicle-to-infrastructure (V2I) scenarios. Additionally, V2X-Seq \cite{yu2023v2x} supports sequential V2X perception by enabling temporal information sharing across agents. However, none of the above datasets provide dense voxel-level semantic annotations required for collaborative 3D semantic occupancy prediction.

\section{Co3SOP Dataset}
\label{sec:cop}
\subsection{Semantic Voxel Annotation Pipeline}
We construct the Co3SOP dataset by replaying multi-agent driving scenarios from OPV2V \cite{9812038} in the CARLA simulator and augmenting them with dense voxel-level semantic occupancy labels. Unlike prior datasets that rely on sparse LiDAR observations, Co3SOP leverages CARLA’s high-fidelity simulation environment to generate complete ground truth.

While CARLA offers a variety of sensors, it does not natively support 3D semantic voxel output. To address this limitation, we develop a custom semantic voxel sensor using Unreal Engine’s built-in collision and overlap detection functions. This sensor efficiently retrieves both occupancy and semantic information at the voxel level, enabling fine-grained annotations for collaborative occupancy prediction tasks.

Dividing a large scene into high-resolution voxels and checking each voxel individually, however, can introduce significant computational overhead. For example, a detection range of $100 \times 100 \times 4.8\,m^3$ at $0.1\,m$ resolution results in approximately $48$ million detection operations for one frame. To alleviate this, we design a multi-stage voxel annotation pipeline, as illustrated in \cref{fig:voxel_sensor} that leverages scene-level object cues to minimize redundant computations:

\textbf{Sensor Configuration.} The voxel is centered at each ego vehicle with specific sensor parameters, including the voxel range and resolution, which are configured to define the spatial extent and granularity. These settings determine how the scene is divided into 3D voxels.

\textbf{Top-Down Broad-Range Box Trace.} After configuration, a top-down overlap detection is performed to coarsely identify all physical objects intersecting with the voxelized scene. Each object’s intersection points (i.e., impact voxels) are recorded and used to initialize a list of seed voxels likely to be occupied. This step refines the focus of annotation to areas likely containing occupied voxels, streamlining further annotation.

\textbf{Object-Level Occupancy Completion.} Then, for each object, a parallel Breadth-First Search (BFS) is initiated from its seed voxels. The BFS selectively propagates to the six axis-aligned neighboring voxels (i.e., $\pm x, \pm y$, and $\pm z$), where each candidate undergoes a collision-aware check for occupancy. Occupied occupied voxels are added to the expansion frontier for subsequent iterations. This localized and iterative process continues until all reachable occupied voxels are exhaustively annotated. The entire design is inherently multi-threaded, with each object assigned to an independent thread, ensuring efficient and scalable occupancy annotation.

\begin{figure*}
    \centering
    \includegraphics[width=0.97\linewidth]{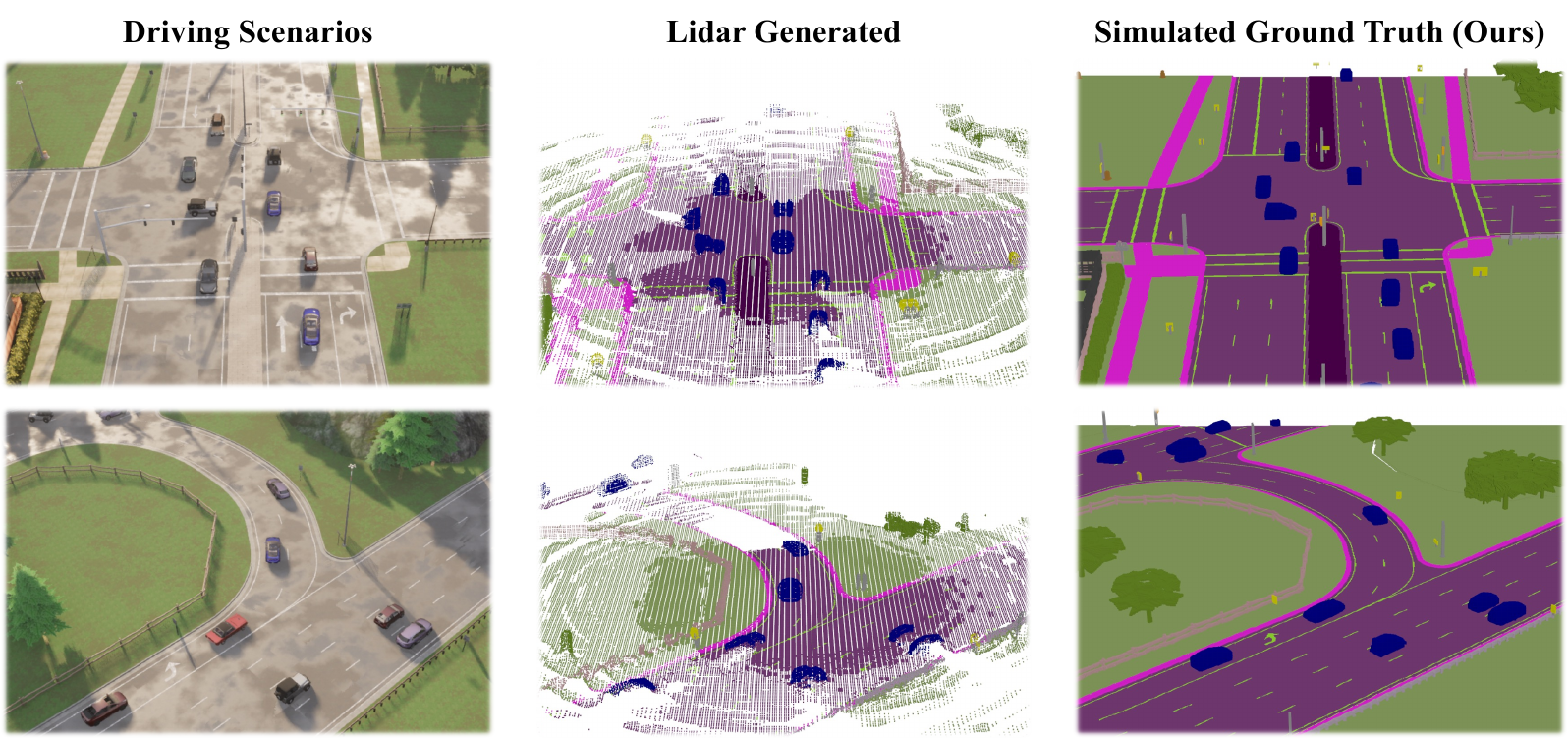}
    \caption{\textbf{Illustration of the V2V scenarios in Carla and the corresponding data collection results.} Left: The screen shot of two V2V scenarios in Carla based on the settings in OPV2V. Mid: LiDAR generated 3D semantic voxel annotations. Right: The annotations collected by our developed 3D semantic voxel sensor.}
    \label{fig:voxels_visual}
\end{figure*}

\textbf{Object Label Assignment.} After occupancy completion, the semantic category of the intersecting object—retrieved from the CARLA simulation engine—is assigned to each of its occupied voxels, completing the voxel-level annotation process.

This optimized annotation pipeline ensures fine-grained spatial annotations by focusing computation on relevant voxel regions, thereby avoiding redundant processing. By decoupling ground-truth generation from sensor visibility, it provides dense and complete voxel labels ideal for collaborative 3D semantic occupancy prediction.

\subsection{Annotation Statistics}
To support tasks with varying ranges and resolutions, we generate dense 3D semantic voxel annotations covering a spatial extent of $100\times100\times7m^{3}$ centered on each ego vehicle. The size of each voxel is $0.1\times0.1\times0.1m^{3}$, resulting in a total voxel resolution of $1000 \times 1000 \times 70$. All voxels are referenced to vehicle’s coordinate, with axes bounded by $x \in [-50,50]m$ (left to right), $y \in [-50,50]m$ (front to back), and $z \in [-2,5]m$ (bottom to top). Additionally, we provide preprocessing tools and the source code for the developed sensor, allowing users to adjust spatial extent or resolution for custom tasks.

To validate the accuracy and completeness of our annotations, we conduct a visual comparison against labels generated from LiDAR scans using the annotation way in SurroundOcc \cite{Wei_2023_ICCV}. \Cref{fig:voxels_visual} illustrates representative V2V scenarios, highlighting voxel labels produced by both approaches. \Cref{fig:voxels_visual} illustrates representative V2V scenarios, highlighting voxel labels produced by both approaches. As illustrated, our pipeline yields significantly denser and more complete semantic voxel annotations, especially in occluded or long-range regions where LiDAR-based annotations suffer from sparsity and missing data.

Co3SOP inherits the multi-agent traffic scenarios of OPV2V \cite{9812038}, retaining its training, validation, and testing splits. Distinctively, we treat each vehicle in a scene as an independent training target, allowing every agent to serve as the ego vehicle while dynamically collaborating with its surrounding vehicles. This formulation not only maximizes data utilization, but also enables flexible many-to-many collaboration configurations across agents. 

Moreover, to enable fine-grained evaluation of collaborative perception, particularly the ability to recover blind spots beyond the ego vehicle’s field of view, we depart from the common practice of marking unobserved voxels as ‘unknown’. Instead, we leverage CARLA’s semantic engine to assign each voxel a category label, including 'empty' for free space, eliminating the unknown labels. As a result, Co3SOP includes 24 semantic categories (including empty), offering broader class diversity than existing 3D semantic occupancy prediction datasets.
\begin{figure*}[htbp]
\centering
\includegraphics[width=0.98\linewidth]{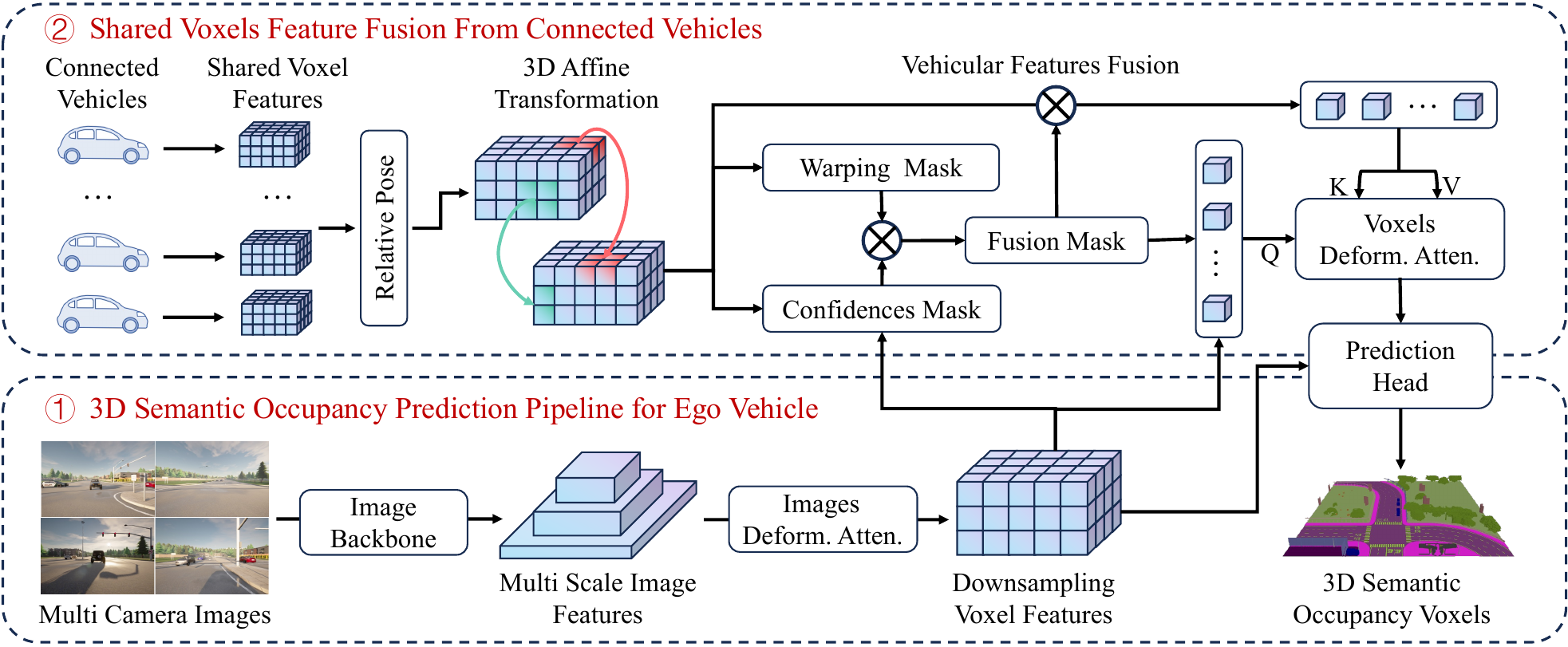}
\caption{The Collaborative 3D Semantic Occupancy Prediction Baseline (Co3SOP-Base) consists of two pipelines: (1) ego prediction pipeline, including image backbone, image deformable attention and prediction head; (2) V2V feature fusion pipeline, including 3D affine transformation and mask voxel deformable attention.}
\label{fig:baseline}
\end{figure*}

\section{Co3SOP Baseline}
To benchmark our proposed Co3SOP dataset, we propose a Collaborative 3D Occupancy Prediction Baseline (Co3SOP-Base), designed to efficiently integrate multi-agent observations and spatial knowledge into unified voxel-based predictions. 
\subsection{Overall Structure}
The overall structure of Co3SOP-Base is illustrated in \cref{fig:baseline}. The framework first uses a shared backbone network to extract multi-scale image features $\{X_{i}^{l}\}_{l=1}^{L}, X_{i}^{l}\in R^{N\times H\times W\times C}$ for each vehicle $V_{i}$, where $L$ represent the scales amount of image features and $i$ is vehicle id. These 2D features are then lifted into the 3D voxel space via an image deformable cross-attention module. Once voxel features $F_{i}\in R^{X\times Y\times Z\times C}$ are obtained for each vehicle, collaborative fusion is enabled by sharing features among connected vehicles (CVs). Upon receiving voxel features from neighboring vehicles, the ego vehicle performs 3D affine transformation to align these features into its own coordinate frame, ensuring proper spatial alignment. Then, the ego vehicle fuses the aligned features with its own voxel representation using a voxel deformable cross-attention mechanism, modulated by a hybrid attention mask that integrates warping and confidence priors. Finally, a prediction head with 3D convolution predicts the voxel-level semantic occupancy map. The details of the key modules are as follows.

\textbf{Image Backbone.} We adopt ResNet101-DCN \cite{he2016deep} as the image backbone to extract multi-scale image features from multi-view camera images. The extracted features are further refined using a Feature Pyramid Network (FPN) \cite{lin2017feature} to aggregate contextual information across scales.

\textbf{Image To Voxel Transformation.} To integrate the 2D image feature into the 3D space, we apply a image deformable cross-attention as in SurroundOcc \cite{Wei_2023_ICCV}. This image deformable cross-attention mechanism aggregates multi-view 2D features and learns to map them into the corresponding 3D voxel locations, by considering geometric relationships, ensuring that spatial consistency is maintained when transitioning from 2D to 3D space:
\begin{align}
    MSDeformAttn(q,p,x) = \nonumber\\ \sum\limits_{m=1}^{N_{head}} W_{m}[\sum\limits_{l=1}^{L}\sum\limits_{k=1}^{K}A_{mlk}&\cdot W_{m}^{'} x_i^l(\phi_l(p_q)+\Delta p_{qmlk})] \nonumber\\
    f_{i} = MSDeformAttn(q_{i}, p_{q}&, \{x_{i}^{l}\}_{l=1}^{L})
\end{align}
where $q_i\in Q_i$ is the corresponding position of query, $Q_i$ is a set of learnable voxel query, $W_{m}$ and $W_{m}^{'}$ are learnable weights, $A_{mlk}$ is attention weight, $p_{q}$ is the reference point of the voxel on the image and $\phi_l$ indicate the rescale sampling function.
    
\textbf{Collaborative Vehicle Spatial Alignment.} Before multi-agent fusion, voxel features from neighboring agents are transformed into the ego vehicle's coordinate frame using a 3D affine transformation module inspired by Spatial Transformer Networks \cite{jaderberg2015spatial}. The transformation matrix is computed from relative poses between agents, ensuring that voxel features from different coordinates are properly aligned for subsequent attention-based fusion:
\begin{align}
    F_{j}^{'} = AffineGrid(F_{j}, T_{ji}^{4\times4})
\end{align}
where $AffineGrid$ is a projective transformation function and $T_{ji}^{4\times4}$ is the pose transformation matrix from vehicle $j$ to vehicle $i$.

\textbf{Confidence Masked Feature Fusion.}
To enable robust and adaptive fusion, each vehicle maintains a learnable confidence mask that estimates the voxel-level reliability of its features. In addition, a binary warping mask is generated during feature alignment to indicate valid projection areas. These two masks are combined to form a hybrid attention mask, which modulates the voxel deformable cross-attention. This design allows the ego vehicle to selectively aggregate features from its collaborators based on both alignment accuracy and confidence, suppressing unreliable or noisy regions. The voxel deformable attention dynamically samples neighboring points across agents, guided by learned offsets and the hybrid attention mask:
\begin{align}
    f_{i}^{'} = MSDeformAttn(f_{i}, p_{ij}, \{f_{j}\}_{j=1}^J)
\end{align}
where $p_{ij}$ is the reference point that $\{f_{j}\}_{j=1}^J$ has higher confidence than $f_{i}$.

\textbf{Prediction Head.} The final stage of Co3SOP-Base is the prediction head, which applies a series of 3D convolutions and 3D deconvolutions to progressively upsample the fused voxel features to the original resolution and output the 3D semantic occupancy results.

\subsection{Loss}
To supervise collaborative 3D occupancy prediction, we combine a voxel-level cross-entropy loss for semantic classification, a scene-class affinity loss \cite{Cao_2022_CVPR} to encourage consistency between semantic and geometric features, and a confidence loss that regularizes the learned confidence mask for inter-agent feature consistency.

\section{Experiments}
\label{sec:experiments}

\subsection{Dataset and Metrics}
To evaluate the performance of collaborative 3D semantic occupancy prediction under varying spatial conditions, we define three perception range settings in our benchmark: $25.6 \times 25.6 \times 4.8\,m^3$ range with $0.1\,m$ voxel size, $51.2 \times 51.2 \times 4.8\,m^3$ range with $0.2\,m$ voxel size, and $76.8 \times 76.8 \times 4.8\,m^3$ range with $0.3\,m$ voxel size. These settings allow us to assess how collaborative perception improves scene understanding as the coverage area increases, particularly in occluded or distant regions where a single vehicle’s sensing capability is limited. For evaluation, we adopt Intersection-over-Union (mIoU), including IoU for each class individually, as well as the mean IoU (mIoU) across all semantic categories.

\begin{table*}[htbp]
\renewcommand\arraystretch{1.08}
\centering
\caption{\textbf{3D semantic occupancy prediction results} on our Co3SOP dataset. We report the mIoU of some state-of-art methods and our baseline model under three tasks with different ranges, i.e., $25.6\times25.6\times4.8 m^3$, $51.2\times51.2\times4.8 m^3$ and $76.8\times76.8\times4.8 m^3$, as well as the IoUs for some object class. The top performances are highlighted in bold.}
\begin{center}
\scalebox{1.18}{
\begin{tabular}{>{\hskip -0.2cm}r<{\hskip -0.1cm}
|  >{\hskip -0.1cm}c <{\hskip -0.3cm} c<{\hskip -0.3cm}  c
<{\hskip -0.1cm}|>{\hskip -0.1cm}c <{\hskip -0.3cm} c<{\hskip -0.3cm}  c
<{\hskip -0.1cm} 
| >{\hskip -0.1cm}c <{\hskip -0.3cm} c<{\hskip -0.3cm}  c
<{\hskip -0.1cm} 
| >{\hskip -0.1cm}c <{\hskip -0.3cm} c<{\hskip -0.3cm}  c
<{\hskip -0.1cm} 
| >{\hskip -0.1cm}c <{\hskip -0.3cm} c<{\hskip -0.3cm}  c
<{\hskip -0.1cm}}

\hline
Method & \multicolumn{3}{c|}{\centering SSCNet}& \multicolumn{3}{c|}{\centering LMSCNet} & \multicolumn{3}{c|}{\centering OccFormer}& \multicolumn{3}{c|}{\centering SurroundOcc}& \multicolumn{3}{c}{\centering \textbf{Co3SOP-Base}}\\
\hline
Modality & \multicolumn{3}{c|}{\centering Lidar}& \multicolumn{3}{c|}{\centering Lidar} & \multicolumn{3}{c|}{\centering Camera}& \multicolumn{3}{c|}{\centering Camera}& \multicolumn{3}{c}{\centering Camera}\\
\hline
Range & 25.6m& 51.2m & 76.8m & 25.6m& 51.2m& 76.8m& 25.6m& 51.2m& 76.8m& 25.6m& 51.2m& 76.8m& 25.6m& 51.2m& 76.8m\\
\hline
mIoU & 13.21 & 9.58 & 10.04 & 24.92 & 20.35 & 17.62 & 29.48 & 25.41 & 24.12 & 28.71 & 25.76 & 24.68& \textbf{30.04} & \textbf{27.50} & \textbf{27.00}\\
\hline
Buildings \raisebox{1\height}{\colorbox{buildings}{}}& 1.84 & 0.17 & 0.19 & 8.67 & 3.09 & 1.79 &\textbf{11.63} & \textbf{11.93} & \textbf{13.43} & 10.63 & 7.57 & 6.91 & 10.05 & 8.19 & 9.15\\
Fences \raisebox{1\height}{\colorbox{fences}{}}& 0.16 & 1.48 & 0.41 & \textbf{22.27} & \textbf{18.01} & 9.26 & 14.17 & 11.60 & 11.04 & 11.06 & 13.28 & 8.84 & 12.37 & 13.70 & \textbf{11.38}\\
Other \raisebox{1\height}{\colorbox{other}{}}& 0.00 & 0.00 & \textbf{16.18} & 0.00 & 0.00 & 0.00 & 0.00 & 0.35 & 2.74& 0.00 & \textbf{1.77} & 12.50 & 0.00 & 0.52 & 6.83\\
Poles \raisebox{1\height}{\colorbox{poles}{}}& 3.60 & 0.14 & 0.00 & \textbf{29.57} & \textbf{24.95} & \textbf{17.92}& 19.67 & 12.62 & 10.17& 17.22& 13.51 &4.78& 20.02 & 16.16&7.12\\
Roadlines \raisebox{1\height}{\colorbox{roadlines}{}}& 0.00 & 0.16 & 0.00 & 2.57 & 0.57 &0.00& \textbf{39.64} & 22.10 &15.53& 26.78 & 22.13 &14.09& 38.43 & \textbf{29.12}& \textbf{16.08} \\
Roads \raisebox{1\height}{\colorbox{roads}{}}& 0.23 & 25.88&0.14& 86.70& 75.84&67.99& 87.40 & 75.30 &73.69& 86.87 & 79.53 &74.87& \textbf{89.24} & \textbf{82.35} &\textbf{80.28}\\
Sidewalks \raisebox{1\height}{\colorbox{sidewalks}{}}& 19.22 & 9.57 &20.28& 42.24 & 48.66 &53.27& 45.32 & \textbf{51.41} & \textbf{59.51}& \textbf{46.61} & 45.23 & 55.30& 46.12 & 42.92&55.47 \\
Vegetation \raisebox{1\height}{\colorbox{vegetation}{}}& 41.43 & 30.89 &22.91& 43.77 & 34.90 &23.91& 42.78 & \textbf{39.77}& \textbf{35.30}& 44.92 & 35.60 &29.08& \textbf{46.36} & 36.30& 34.26 \\
Vehicles \raisebox{1\height}{\colorbox{vehicles}{}}& 71.73 & 48.09 &39.35& \textbf{85.35} & \textbf{75.63} &\textbf{62.94}& 75.70 & 51.25 &33.35& 75.95 & 52.34&29.20& 80.55 & 66.48 &50.98\\
Walls \raisebox{1\height}{\colorbox{walls}{}}& 0.26 & 0.49 &0.18& 9.97 & 10.39 &10.08& \textbf{13.41} & \textbf{15.53} &11.55& 12.37 & 12.92 & 10.03&12.11 & 13.99& \textbf{12.70}\\
Trafficsigns \raisebox{1\height}{\colorbox{trafficsigns}{}}& 0.00 & 0.00 &0.00& \textbf{18.19} & 0.02 &0.04& 9.73 & 7.68 &2.49& 17.27 & \textbf{11.72} &6.23& 11.16 & 9.16 &\textbf{10.02}\\
Ground \raisebox{1\height}{\colorbox{ground}{}}& 37.73 & 0.08 & 22.18& 62.68 & 31.81 & 20.59& \textbf{67.08} & \textbf{57.79}& 64.75 & 53.80 & 52.90 &64.77& 55.84 & 51.96 &\textbf{68.88}\\
Bridge \raisebox{1\height}{\colorbox{bridge}{}}& 0.00 & 0.03 & 0.07& 0.00 & 0.00 &0.00& 0.00 & \textbf{2.95}&\textbf{5.45} & 0.00 & 2.32 & 3.74& 0.00 & 2.26&4.39\\
Guardrail \raisebox{1\height}{\colorbox{guardrail}{}}& 8.22 & 12.72& 10.21& 12.02 & 6.07 &3.37& 35.53 & 41.41 &36.90& 48.49 & 42.17 &39.65& \textbf{53.23} & \textbf{48.54}& \textbf{44.89} \\
Trafficlight \raisebox{1\height}{\colorbox{trafficlight}{}}& 0.25 & 0.00 &0.00& 0.00 & 0.00 &0.00 & \textbf{5.43} & \textbf{3.75} & 0.11 & 2.11 & 2.03 & 1.10& 1.27 & 3.30 &\textbf{2.42} \\
Terrain \raisebox{1\height}{\colorbox{terrain}{}}& 26.41 & 2.74 &17.01 &36.11&36.93&33.43&\textbf{86.95}&53.91&53.05&76.58&75.08&\textbf{75.08}&82.93&\textbf{80.75}&74.43 \\
\hline
\end{tabular}
}
\end{center}
\label{table:all}
\end{table*}



\subsection{Benchmark Methods}
To establish a fair comparison under our benchmark, we evaluate four representative 3D semantic occupancy prediction models from two modalities. For LiDAR-based methods, we select SSCNet \cite{song2017semantic} and LMSCNet \cite{9320442}; for camera-based methods, we include SurroundOcc \cite{Wei_2023_ICCV} and OccFormer \cite{Zhang_2023_ICCV}. These models are chosen for their widespread adoption and strong performance in prior benchmarks. We adopt their official implementations and default configurations, modifying only the data loaders and batch sizes to ensure compatibility with the Co3SOP dataset.

\subsection{Implementation Details}
For the image backbone, we adapted the weights from FCOS3D \cite{wang2021fcos3d} as the pretrained weights and set the number of output levels to 4. And these image features from stages 1, 2, and 3 are then fed into the FPN to obtain 4 levels of multi-scale features. For the image and voxel deformable attention, we set the number of layers as 3 and 1. For intermediate voxel features, we apply a downsampling rate of $\frac{1}{4}$ and upsample them in the prediction head. For the collaborative setting, we set the maximum number of collaborating agents to 6 per ego vehicle, based on number of vehicles in OPV2V. During the training process, we apply multi-scale supervision for the multi-scale outputs from prediction head and image augmentation as in SurroundOcc \cite{Wei_2023_ICCV}. All experiments are conducted on 8 RTX 4090 GPUs.

\subsection{Benchmark Analysis}
The performance of the benchmark methods, SSCNet, LMSCNet, SurroundOcc, and OccFormer, along with our proposed Co3SOP-Base, is presented in \cref{table:all}, which reports mean IoU (mIoU) as well as class-wise IoUs across three perception ranges.

Among the four selected methods, OccFormer achieves the highest mIoU in the $25.6,m$ range (29.48), while SurroundOcc performs best in the $51.2,m$ and $76.8,m$ ranges (25.76 and 24.68, respectively). Notably, both camera-based methods consistently outperform LiDAR-based methods in overall mIoU. However, LiDAR-based models still demonstrate strengths in certain object classes particularly in small-range such as Vehicles and Poles. 

Compared with the four selected methods, our proposed Co3SOP-Base consistently achieves the best overall performance across all three perception ranges, with mIoUs of 30.04 ($25.6m$), 27.50 ($51.2m$), and 27.00 ($76.8m$). Compared to the single vehicle methods, our proposed baseline gaines greater improvements as the prediction range increases, especially for object classes such as vehicles, guardrail and roadlines. These results highlight the effectiveness of collaborative voxel fusion, especially as the spatial extent increases.

\begin{table*}[htbp]
\renewcommand\arraystretch{1.2}
\setlength{\abovecaptionskip}{0cm}
\centering
\caption{Ablation study on the impact of collaboration in 3D semantic occupancy prediction.}
\begin{center}
\scalebox{1.18}{
\begin{tabular}{c | c | c@{\hspace{0.1cm}} c@{\hspace{0.1cm}} c@{\hspace{0.1cm}} c@{\hspace{0.1cm}} c@{\hspace{0.1cm}} c@{\hspace{0.2cm}} c@{\hspace{0.1cm}} c@{\hspace{0.1cm}} c@{\hspace{0.1cm}} c@{\hspace{0.1cm}}c@{\hspace{0.1cm}}c@{\hspace{0.1cm}}c@{\hspace{0.1cm}}c@{\hspace{0.1cm}}c@{\hspace{0.1cm}}c@{\hspace{0.1cm}} | c}
\hline
\multicolumn{1}{c|}{Range}
&\multicolumn{1}{c|}{\makecell*[b]{\rotatebox{90}{Collaboration}}}
& \makecell*[b]{\rotatebox{90}{\raisebox{1\height}{\colorbox{buildings}{}}\,Buildings}}
& \makecell*[b]{\rotatebox{90}{\raisebox{1\height}{\colorbox{fences}{}}\,Fences}}
& \makecell*[b]{\rotatebox{90}{\raisebox{1\height}{\colorbox{other}{}}\,Other}}
& \makecell*[b]{\rotatebox{90}{\raisebox{1\height}{\colorbox{poles}{}}\,Poles}}
& \makecell*[b]{\rotatebox{90}{\raisebox{1\height}{\colorbox{roadlines}{}}\,Roadlines}}
& \makecell*[b]{\rotatebox{90}{\raisebox{1\height}{\colorbox{roads}{}}Roads}}
& \makecell*[b]{\rotatebox{90}{\raisebox{1\height}{\colorbox{sidewalks}{}}Sidewalks}}
& \makecell*[b]{\rotatebox{90}{\raisebox{1\height}{\colorbox{vegetation}{}}\,Vegetation}}
& \makecell*[b]{\rotatebox{90}{\raisebox{1\height}{\colorbox{vehicles}{}}\,Vehicles}}
& \makecell*[b]{\rotatebox{90}{\raisebox{1\height}{\colorbox{walls}{}}\,Walls}}
& \makecell*[b]{\rotatebox{90}{\raisebox{1\height}{\colorbox{trafficsigns}{}}\,Trafficsigns}}
& \makecell*[b]{\rotatebox{90}{\raisebox{1\height}{\colorbox{ground}{}}\,Ground}}
& \makecell*[b]{\rotatebox{90}{\raisebox{1\height}{\colorbox{bridge}{}}\,Bridge}}
& \makecell*[b]{\rotatebox{90}{\raisebox{1\height}{\colorbox{guardrail}{}}\,Guardrail}}
& \makecell*[b]{\rotatebox{90}{\raisebox{1\height}{\colorbox{trafficlight}{}}\,Trafficlight}}
& \makecell*[b]{\rotatebox{90}{\raisebox{1\height}{\colorbox{terrain}{}}\,Terrain}}
& \multicolumn{1}{|c}{mIoU}\\
\hline

\multirow{2}{*}{$25.6m$} &$\times$ &9.96&12.56&0.00&18.73&36.19 &88.53&44.69&45.51&77.53&11.13 &11.10&55.08&0.00&48.61&1.50&82.49&\multicolumn{1}{|c}{29.36} \\
&$\checkmark$&10.05&12.37&0.00&20.02&38.43 &89.24 &46.12&46.36&80.55 &12.11 &11.16&55.84&0.00&53.23&1.27&82.93&\multicolumn{1}{|c}{30.04} \\
\hline
\multirow{2}{*}{$51.2m$} &$\times$&7.70&12.60&2.54&10.34&26.17&80.03&41.73&32.79&54.78&13.25&8.53&46.76&2.23&42.83&3.35&78.87&\multicolumn{1}{|c}{25.69} \\
&$\checkmark$&8.19&13.70&0.52&16.16&29.12 &82.35&42.92 &36.30&66.48 &13.99 &9.16&51.96&2.26&48.54&3.30&80.75&\multicolumn{1}{|c}{27.50}  \\
\hline
\multirow{2}{*}{$76.8m$} &$\times$&7.32&9.88&12.04&4.29&14.55 &75.54&53.53&31.18&34.32&10.54&7.41&62.58&4.02& 40.02&2.18&71.54&\multicolumn{1}{|c}{24.81} \\
&$\checkmark$&9.15&11.38&6.83&7.12&16.08 &80.28 &55.47 &34.26 &50.98&12.70&10.02&58.88&4.39&44.89&2.42&74.43&\multicolumn{1}{|c}{27.00}  \\
\hline
\end{tabular}
}

\end{center}
\label{table:ablation}
\end{table*}

\subsection{Ablation Study}
\textbf{Effect of Collaborative Feature Fusion.} We investigate the impact of inter-agent collaboration by comparing performance with and without V2V feature fusion across all three perception ranges. As shown in \cref{table:ablation}, collaboration consistently improves mIoU across three ranges. The gains are most significant in the large-range setting, where agents have broader visibility overlap and benefit more from complementary viewpoints. Object-level analysis reveals that collaboration notably enhances performance for large or occlusion-prone classes such as Vehicles (e.g., $34.43 \xrightarrow{}50.98$ IoU at $76.8m$) and Poles (e.g., $10.34 \xrightarrow{}16.16$ IoU at $51.2m$). However, for thin or rare classes like Trafficlight, collaboration yields little to no improvement—and in some cases even slight drops (e.g., Trafficlight: $3.35\xrightarrow{}3.30$ at $51.2m$).

\begin{figure}[h]
    \centering
    \includegraphics[width=1.\linewidth]{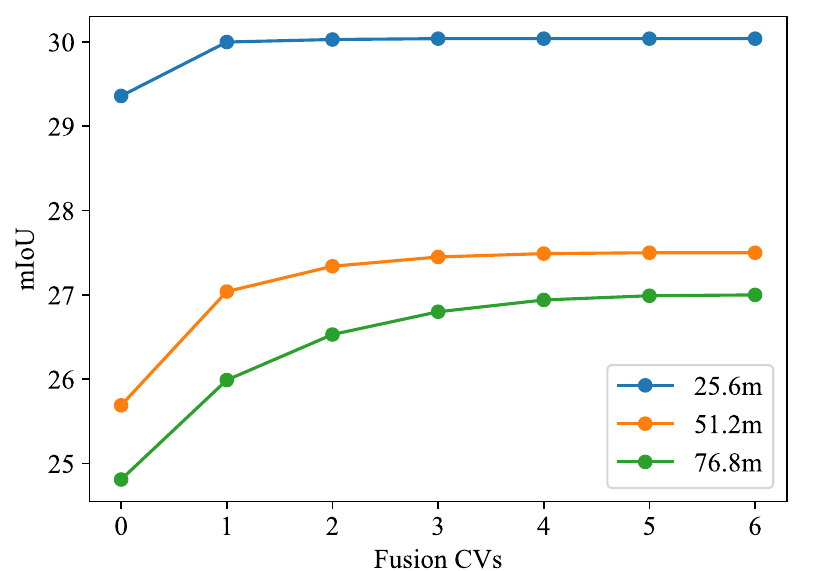}
    \captionof{figure}{Ablation study on the number of collaborating vehicles used for feature fusion.}
    \label{fig:num_collaborative}
\end{figure}

To further explore the effect of collaboration scale, we evaluate performance under varying numbers of collaborating vehicles. Collaborating vehicles are selected based on spatial proximity, sorted by distance to the ego vehicle. As shown in \cref{fig:num_collaborative}, performance gains from collaboration mainly come from 1–2 nearby agents. At shorter ranges ($25.6m$), only the closest collaborator yields improvement, while larger-range settings ($51.2m$, $76.8m$) benefit from additional agents due to increased spatial coverage and complementary viewpoints.

\textbf{Robustness on Pose Noise}
We further assess the robustness of Co3SOP-Base under localization uncertainty by injecting Gaussian pose noise into the relative transformations used for inter-agent feature alignment. As shown in \cref{tab:posenoise}, increasing the mean offset from $\mu=0.1m$ to $0.5m$ (with fixed standard deviation $\sigma=0.02$) leads to a gradual mIoU degradation across all ranges. This decline illustrates that misalignment in inter-agent transformations can affect collaborative feature quality. However, the performance remains relatively stable under moderate noise, suggesting that our warping-based spatial alignment and confidence-aware fusion retain reasonable robustness in the presence of pose perturbations.



\begin{table}[h]
\renewcommand\arraystretch{1.18}
    \centering
    \caption{ Ablation study on the impact of pose noise on collaborative 3D semantic occupancy prediction.}
    \scalebox{1.18}{
    \begin{tabular}{c|c | c| c}
    \hline
        &\multicolumn{3}{c}{Range}\\
        \cline{2-4}
         Noise & 25.6m& 51.2m& 76.8m \\
         \hline
         $\mu=0.0, \, \sigma=0.00$& 30.04& 27.50&27.00\\
         $\mu=0.1, \, \sigma=0.02$& 29.95& 27.38&26.96\\
         $\mu=0.2,\, \sigma=0.02$& 29.78& 27.13&26.83\\
         $\mu=0.3,\, \sigma=0.02$& 29.59& 26.82&26.64\\
         $\mu=0.4,\, \sigma=0.02$& 29.41& 26.50&26.39\\
         $\mu=0.5,\, \sigma=0.02$& 29.30& 26.20&26.15\\
    \hline
    \end{tabular}
    }
    
    \label{tab:posenoise}
\end{table}

\section{Conclusion}
We presented Co3SOP, a synthetic benchmark for collaborative 3D semantic occupancy prediction, built upon high-fidelity simulation and comprehensive voxel annotations. Unlike prior datasets which are limited by LiDAR sparsity, Co3SOP leverages a custom-designed voxel annotation pipeline to provide complete and fine-grained semantic voxel labels, enabling robust evaluation of multi-agent perception systems. We further proposed Co3SOP-Base, a baseline framework incorporating confidence and alignment aware masked voxel deformable attention for multi-agent feature fusion. Extensive experiments demonstrate the effectiveness of V2V collaboration in improving perception performance. This work addresses the critical gap in collaborative 3D voxel-level understanding by providing both a benchmark and a baseline, paving the way for safer and more robust autonomous driving systems.










\bibliographystyle{IEEEtran}
\bibliography{root}
\addtolength{\textheight}{-12cm}   

\end{document}